\documentclass[runningheads]{llncs}
\usepackage{graphicx}

\usepackage{tikz}
\usepackage{comment}
\usepackage{amsmath,amssymb} 
\usepackage{color}

\usepackage{booktabs}
\usepackage{multirow}

\usepackage{float}
\usepackage{algorithm}
\usepackage{algpseudocode}

\usepackage[accsupp]{axessibility}  

\begin{document}
\pagestyle{headings}
\mainmatter
\def\ECCVSubNumber{1956}  

\title{Gait Recognition with Mask-based Regularization} 

\titlerunning{Gait Recognition with Mask-based Regularization}
%
\author{Chuanfu Shen\inst{1,2}\textsuperscript{$\dagger$} \and
Beibei Lin\inst{3}\textsuperscript{$\dagger$} \and
Shunli Zhang\inst{3} \and\\
George Q. Huang\inst{1} \and 
Shiqi Yu\inst{2}\thanks{Corresponding author.\\ $\dagger$ C.S. and B.L. are co-first authors.} \and
Xin Yu\inst{4} 
}
\authorrunning{C. Shen et al.}
%
\institute{The University of Hong Kong \and 
Southern University of Science and Technology \and
Beijing Jiaotong University \and
University of Technology Sydney\\
} 
\maketitle

\begin{abstract}

      Most gait recognition methods exploit spatial-temporal representations from static appearances and dynamic walking patterns. 
      However, we observe that many part-based methods neglect representations at boundaries. In addition, the phenomenon of overfitting on training data is relatively common in gait recognition, which is perhaps due to insufficient data and low-informative gait silhouettes.
      Motivated by these observations, we propose a novel mask-based regularization method named ReverseMask. By injecting perturbation on the feature map, the proposed regularization method helps convolutional architecture learn the discriminative representations and enhances generalization. Also, we design an Inception-like ReverseMask Block, which has three branches composed of a global branch, a feature dropping branch, and a feature scaling branch. Precisely, the dropping branch can extract fine-grained representations when partial activations are zero-outed. Meanwhile, the scaling branch randomly scales the feature map, keeping structural information of activations and preventing overfitting. The plug-and-play Inception-like ReverseMask block is simple and effective to generalize networks, and it also improves the performance of many state-of-the-art methods. Extensive experiments demonstrate that the ReverseMask regularization help baseline achieves higher accuracy and better generalization. Moreover, the baseline with Inception-like Block significantly outperforms state-of-the-art methods on the two most popular datasets, CASIA-B and OUMVLP. The source code will be released.
      
\end{abstract}

\section{Introduction}
\label{sec:intro}

    Gait recognition~\cite{nixon2010humanbook} utilizes appearance and walking patterns as clues to identify people from images sequences. Gait recognition can achieve perception-free human identification at a distance, which is hardly achievable by other biometrics such as the face, fingerprint, iris. Nevertheless, gait recognition is still facing several challenges such as pose\cite{ptsn}, carrying and clothing\cite{casiab}, ageing\cite{isir}, illumination, and occlusions.
    
\begin{figure}[ht!]
\centering
\includegraphics[width=0.60\linewidth]{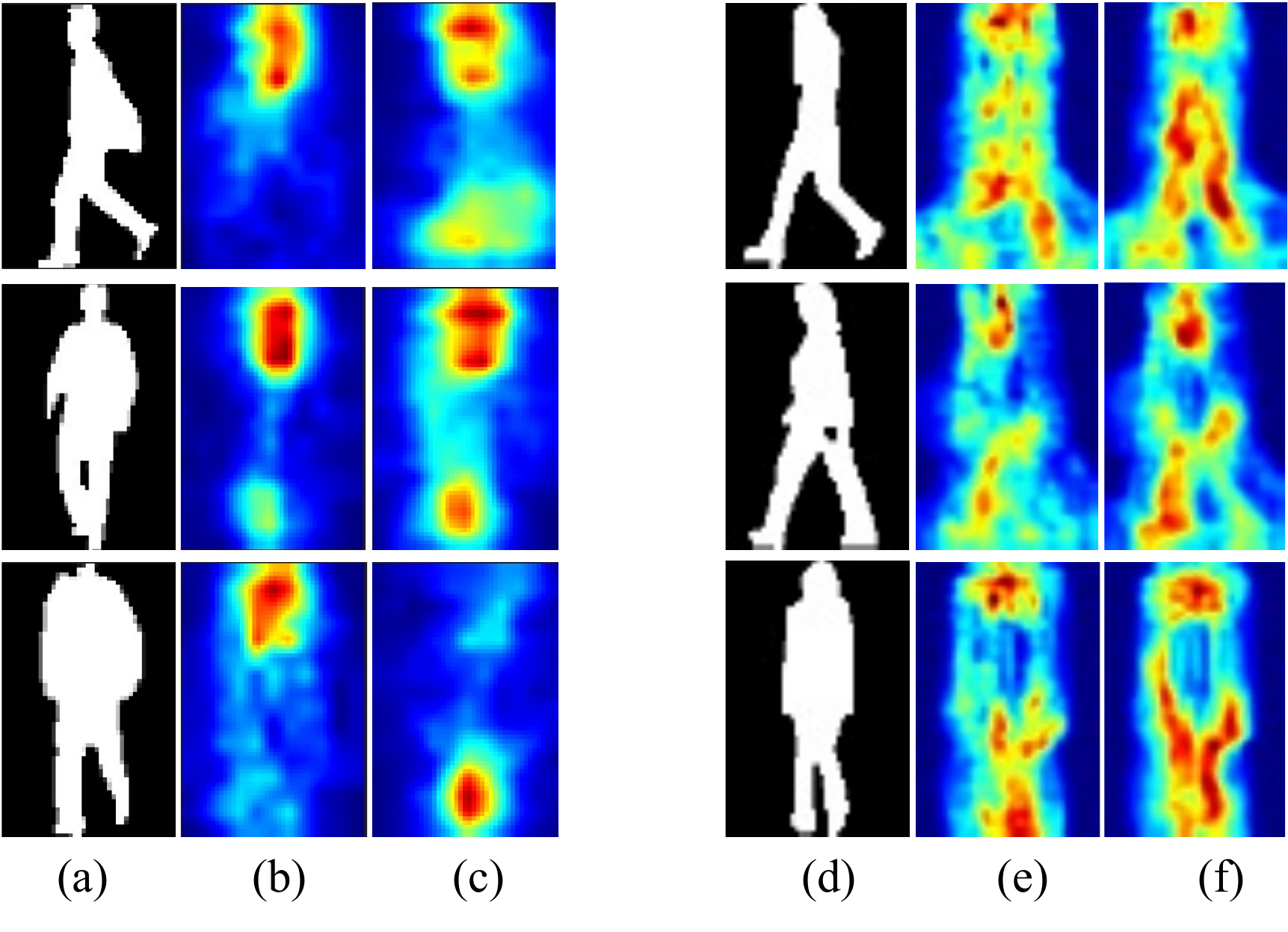}
\centering
\caption{
Visualization~\cite{visualmethod} of activation for different methods on CASIA-B. \textbf{Left}: (a) input silhouettes. (b) activation visualization of GaitSet~\cite{aaai2019gaitset}. (c) GaitSet with our proposed ReverseMask regularization method. \textbf{Right}: (d) input silhouettes. (e) activation visualization of GaitGL~\cite{gaitgl}. (f) GaitGL with ReverseMask regularization, respectively.
GaitSet concentrates on the head and foot regions. GaitGL distributes the attention on different moving parts of a human body, but the noncontinuous attention reveals that information is missing at the horizontal boundaries. }
\label{fig:heatmap_firstpage}
\vspace{-5mm}
\end{figure}

    Recently, several studies~\cite{shiraga2016geinet,wu2016comprehensive,20163dcnn,2019_part_zhangyuqi,MT3D} achieved impressive performance to facilitate gait recognition using the deep convolutional neural networks. Notably, most appearance-based gait recognition models only adopt a limited number of stacked layers (refers to depth). In contrast, other visual recognition tasks~\cite{action,he2021masked} have already greatly benefited from very deep models. However, such deep architecture is inferior to shadow networks in gait recognition. To our best knowledge, the preference to apply shallow models is mainly caused by two aspects: i) The task of gait recognition has less data for training. For example, Sports-1M and YouTube-8M~\cite{sports1m,youtube} contain millions of action videos, while the largest cross-view gait dataset~\cite{oumvlp} provides 0.1 million gait sequences for training. ii) Silhouettes provide less information than information from the RGB modality. Therefore, we think that insufficient training data can easily lead many deep networks~\cite{aaai2019gaitset,gln,disentangled2019cvprgaitnet} to risk overfitting the salient characteristics. The \textit{overfitting} phenomenon can be glimpsed from the activations visualization of classical GaitSet\cite{aaai2019gaitset} as shown in Fig.~\ref{fig:heatmap_firstpage} (b). The learned representations easily focus on the most discriminative pattern, but it leads to poor generalization performance on validation. Unexpectedly, this problem has not been indicated well yet in the previous literature.
    

    By partitioning the holistic human body into many horizontal parts, the part-based methods~\cite{gaitpart,gaitgl,2019_part_zhangyuqi} leverage partial features to prevent the issue of \textit{overfitting} on salient representations.
    Nonetheless, the learned representations by part-based methods are noncontinuous and distributed sparsely as shown in Fig.~\ref{fig:heatmap_firstpage} (e). The conventional part-based methods employ hard boundary partition, where spatial clues of each part can only concentrate on inner partial regions, neglecting inter-part correlation. This phenomenon of noncontinuous representations refers to \textit{boundary} isolation in this paper. Therefore, it is necessary to prevent \textit{overfitting} and \textit{boundary}, furthermore improving the generalization ability and performance of deep networks for robust gait recognition.

    To address the \textit{overfitting} issue, various data augmentation and regularization methods~\cite{dropout,batchdropblock} have been proposed, such as input-level random erasing~\cite{RandomErasing} and feature-level DropBlock~\cite{ghiasi2018dropblock}. 
    The principle of these methods is to inject noise into raw input or feature, producing extra data so that convolutional networks do not overfit the training data. We argue that the main drawback of various erasing-based methods is that it only zero-out features. Besides, it also can apply perturbation by scaling activations. The scaling features extents to bring about more noise to prevent network overfitting. Moreover, it is also perfectly suitable for appearance-based gait recognition because the scaling regularization supervises the network to look for structural evidence of gait for simply silhouettes.
    For the \textit{boundary} issue, we analyze this isolation is mainly caused by manual partition like GaitPart~\cite{gaitpart}, where such convolutional layers can only capture internal representations but ignore semantic information at the boundary. Therefore, it is straightforward to consider generating random partition to avoid neglecting regional representations during training to overcome the \textit{boundary} issue. 

    The aforementioned intuitions inspires us to address both \textit{overfitting} and \textit{boundary} problems by introducing a mask-based regularization method. In this work, we propose a novel regularization method called ReverseMask, with a corresponding Inception-like ReverseMask Block. Specifically, the Inception-like ReverseMask Block has a parallel architecture consisting of three branches, which are global branch, dropping branch, and scaling branch. Within both dropping branch and scaling branch, the novel ReverseMask layer is plugging as a regularizer where receiving features from the previous layer and then producing a pair of features with perturbation. Specifically, the ReverseMask layer zero-outs partial features for the dropping branch. Therefore, the dropping branch effectively captures fine-grained representations since convolutional filters must look at informative regions to fit the erased feature. For the scaling branch, ReverseMask randomly scales the value of activations so that the perturbation forces a convolutional filter to learn the structural characteristics of gait silhouettes.
    
    In our experiments, adding Inception-like ReverseMask Block to GaitSet shows its regularizing convolutional networks as shown in Fig.~\ref{fig:heatmap_firstpage} as well as improving generalization performance in cross-view gait recognition under clothing from 74.8\% to 76.3\%. Besides, Inception-like ReverseMask Block relieves the phenomena of \textit{boundary} isolation of part-based methods and improves cross-view gait recognition accuracy as well.
    In summary, the contributions of this work are listed as follows:
    \begin{itemize}
        
        \item The novel ReverseMask layer is superior to regular DropBlock regularization in our experiments. The ReverseMask provides much more stable regularization and speeds up the training.
    
        \item We propose a novel Inception-like ReverseMask Block with the scaling branch, which helps to capture structural gait representations. In particular, Inception-like ReverseMask Block can be flexibly embedded into the most recent gait frameworks, improving the discriminativeness of feature representations and the generalization performance of models.
        
        \item The network with proposed regularization method outperforms start-of-the-art methods on two popular benchmarks: CASIA-B~\cite{casiab} and OUMVLP~\cite{oumvlp}.
    \end{itemize}

\section{Related Work}

\subsection{Gait Recognition}
    
    \subsubsection{Holistic-based recognition.} To extract holistic gait features, template-based methods~\cite{shiraga2016geinet,wu2016comprehensive} utilize convolutional neural networks directly on gait templates like Gait Energy Images~\cite{han2005gei} (GEI) in early. Wu \etal~\cite{wu2016comprehensive} propose three types of architecture to recognize the most discriminative changes of gait features and provide many comprehensive experiments on cross-view gait recognition performance. Nevertheless, template-based methods lose temporal and fine-grained spatial information. In contrast, many sequence-based methods~\cite{aaai2019gaitset,gln,pami2020gaitnet} conduct feature extractors on each frame to capture detailed spatial clues across frames. For example, Chao \etal~\cite{aaai2019gaitset} regard a gait as a set of independent frames, then aggregate and fuse the extracted set-level feature by the proposed set pooling unit. However, the gait recognition methods based on holistic representations tend to focus on the most representative salient patterns, leading to distinguishing the subjects trickily.
    
    \subsubsection{Part-based recognition.} The part-based methods~\cite{gaitpart,gaitgl,2019_part_zhangyuqi} extensively exploit fine-grained spatial cues from multiple parts for local representation learning. The part-based methods can extract from different types of local regions \ie patch~\cite{patch}, body components~\cite{usflabel,ptsn}, vertical or horizontal bins~\cite{gaitmaskold}, and attentive regions~\cite{reid_attention_part}. GaitPart~\cite{gaitpart} introduces the Focal Convolution Layer to enhance the part-level spatial features by splitting the input feature map into several parts horizontally. Horizontal Pooling is widely used in many approaches~\cite{gln,cstl,gaitgl} since GaitSet~\cite{aaai2019gaitset} adopts the Horizontal Pyramid Pooling~\cite{HPP} from person re-identification. Specifically, Horizontal Pyramid Pooling separates feature maps into hierarchical strips with multi-scales hierarchy, improving spatial discriminative ability. Most works\cite{gaitgl,gaitpart,2019_part_zhangyuqi} are based on pre-defined uniform partition leading to the problem of boundary. Zhang \etal~\cite{2019_part_zhangyuqi} propose a method based on the learned partition, but boundary isolation still exists. Recently, GaitGL~\cite{gaitgl} combines both global and partial features, gaining more robust spatial representation. Besides, it also introduces 3D ConvNets to learn spatial-temporal integrated features for gait recognition. However, no matter learned or hand-crafted partition-based methods have introduced the boundary problem, neglecting the spatial information at the foundation.
\begin{figure}[ht!]
\centering
\includegraphics[width=0.8\linewidth]{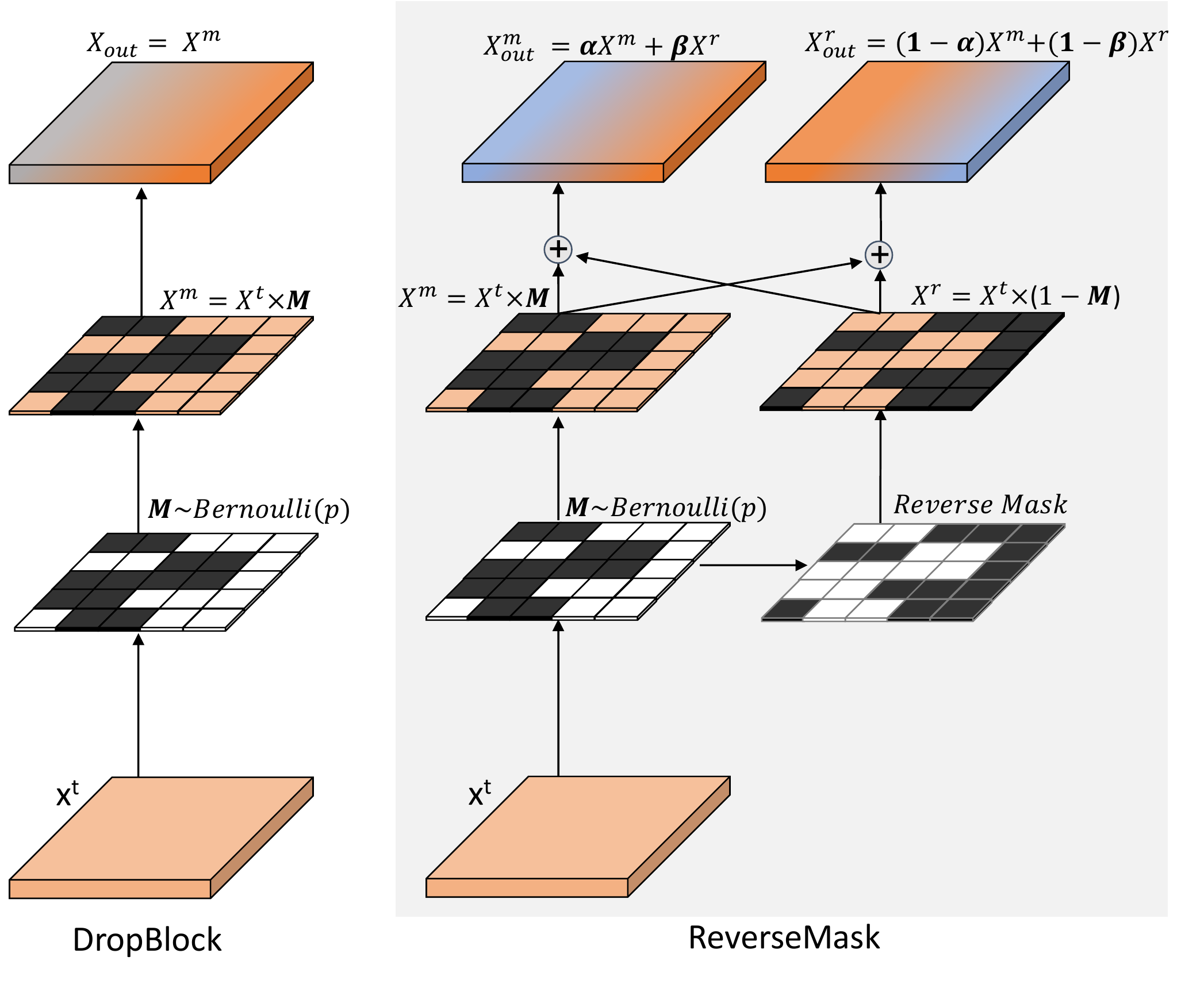}
\centering
\caption{
Illustration of our proposed ReverseMask. Given two random variable, ReverseMask introduces perturbation on networks by reducing values of feature map on selected regions.}
\label{fig:reversmask}
\vspace{-5mm}
\end{figure}
\subsection{Erasing Images or Activations}

    \subsubsection{Image based erasing}~\cite{cutout,RandomErasing} is widely adopted as a data augmentation technique. In recent years, cutout~\cite{cutout} has been demonstrated that masking out partial feature maps can improve generalization of convolutional neural networks and achieve better performance in many tasks such as object detection~\cite{bochkovskiy2020yolov4} and person re-identification~\cite{RandomErasing}.

    \subsubsection{Feature based erasing}~\cite{batchdropblock,ghiasi2018dropblock} is an alternative regularization technique that is implemented by using zero-masking directly on the feature map. Dropout~\cite{dropout} is effective to prevent overfitting, but it designs initially for fully-connected layer. While the mechanism of dropout also brings many successful works on convolutional neural networks, such as SpatialDropout~\cite{spatialdropout} and DropBlock~\cite{ghiasi2018dropblock}.

    Our method drops identical, randomly selected regions of convolutional features for sequences in a batch, which has been proved effective in previous literature~\cite{batchdropblock}. However, our work presents a scaling mechanism where parts of convolutional features are multiplied by a random ratio~\cite{beyonddropout}. Feature maps are multiplied by a random ratio rather than zero, presenting novel structural representations for robust gait recognition. Furthermore, we also argue that such structural representations are ignored, while many methods~\cite{gaitpart,gaitmask} only proposed to capture local representations over coarse pre-defined parts\cite{gaitgl,gaitmaskold}.

\begin{figure}[ht]
\centering
\includegraphics[width=0.95\linewidth]{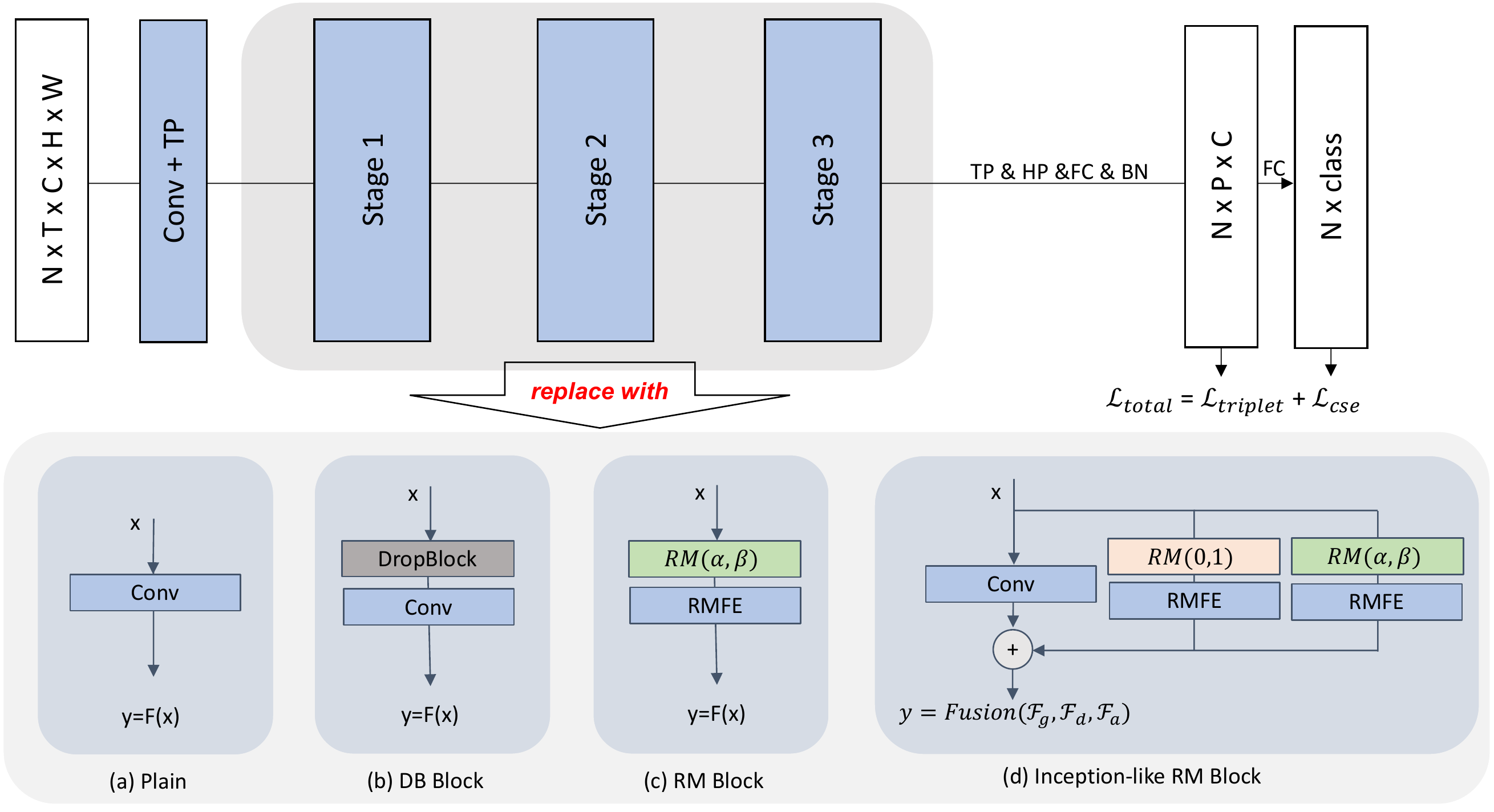}
\centering
\caption{
The framework of our proposed method. \textbf{Top}: a simplified model~\cite{gaitgl} as our baseline model; \textbf{Bottom}: Four replaceable blocks. \textit{TP, FC, HP} and \textit{BN} mean temporal pooling aggregation layer~\cite{gaitgl,aaai2019gaitset}, fully connected layer, horizontal pooling layer\cite{HPP}, and batch normalization layer~\cite{batchnorm}, respectively. \textit{Conv} indicates a convolutional layer followed by \textit{Leaky ReLU}~\cite{leakyrelu}. \textit{\textbf{RM}} is our proposed \textit{ReverseMask} layer, and it receive a pair variable to inject noise on feature map. While \textbf{\textit{RMFE}} is the feature extraction layer specifically designed for \textit{ReverseMask}.
}
\label{fig:ratio}
\vspace{-5mm}
\end{figure}

\section{Our approach}
In this work, we propose a novel regularization method named ReverseMask, which can help relieve convolutional architecture overfitting on training sets. In this section, we first formulate the simple yet effective ReverseMask (Sec.~\ref{subsection:revermask}). Then the detailed definitions of other corresponding components are followed (Sec.~\ref{subsection:fe}). Finally, two variants of building block are introduced (Sec.~\ref{subsection:block}) for making ReverseMask as easy-to-use as possible.

\subsection{ReverseMask Layer}
\label{subsection:revermask}

    ReverseMask is a simple regularization method similar to DropBlock. Its main difference from DropBlock is significant. (1) ReverseMask is a general mask-based regularizer, which produces a pair of masked features instead of only one masked feature. (2) ReverseMask introduces perturbation on networks by reducing values of feature maps on selected regions. Illustrations of ReverseMask and DropBlock are shown in Fig.~\ref{fig:reversmask}. Notably, there are only two parameters for ReverseMask, which are {$reg\_prob$} and $p$. $reg\_prob$ determines the probability of changing activations undergoing ReverseMask. \textit{p} controls the area ratio of Masked generation. 
    
    During the training stage, we first randomly sample mask on the given feature map $X_l\in\mathbb{R}^{ b_l \times n_l \times c_l \times h_l \times w_l}$ at the $l$th layer. Then its paired mask can be determined by reversing the mask matrix. The process can be defined as
        \begin{equation}
            \label{eq:reversemask}
         \mathit M^{m}_{i,j} \sim  Bernoulli(p)
        \end{equation}
        \begin{equation}
        \label{eq:reversemask_p}
         \mathit M^{p}_{i,j} = 1 - M^{m}_{i,j} 
        \end{equation}
    where $M^m\in\mathbb{R}^{ h_l \times w_l}$ and $M^p\in\mathbb{R}^{ h_l \times w_l}$ refer to sampled mask and its paired mask. Notably, such a pair of masks is temporally synchronized, and we only conduct experiments with shared masks across different feature channels for simplicity and fair comparison with part-based methods. Then, applying the paired of reversal masks on given features, two masked features can be obtained
        \begin{equation}
         \mathit X^m_l = X_l \times M^m
        \end{equation}
        \begin{equation}
         \mathit X^r_l = X_l \times M^r
        \end{equation}
    It is worth noticing that these masked features are identical to features produced by DropBlock, but ReverseMask has a pair of masked features. Finally, we introduce perturbation on activations for regularization with a pair of randomly generated values
            \begin{equation}
         \mathit \alpha,\beta \sim  Uniform(0,1)
        \end{equation}
        \begin{equation}
                    \label{eq:featmask}
         \mathit X^{m}_{l_{out}} = \alpha X^m_l + \beta X^r_l
        \end{equation}
        \begin{equation}
                     \label{eq:featpaired}
         \mathit X^{p}_{l_{out}} = (1-\alpha) X^m_l + (1-\beta) X^r_l
        \end{equation}
    where two random variables $\alpha$ and $\beta$ satisfy the uniform distribution to perturb activations. $X^{m}_{l_{out}}\in\mathbb{R}^{ b_l \times n_l \times c_l \times h_l \times w_l}$ and $X^{p}_{l_{out}}\in\mathbb{R}^{ b_l \times n_l \times c_l \times h_l \times w_l}$ are obtained as input of further feature learning.
    Therefore, our proposed ReverseMask is a more general DropBlock, generating zero-masked features and scaling features to regularize the network. Beyond feature-level erasing methods, our proposed mask-based scaling regularization method is able to capture structural information from gait silhouettes. At the same time, erasing-based regularizer tends to perform better on fine-grained representation learning.

\subsubsection{Mask Sampling.} In the experiments, we have studied many mask sampling strategies, such as masking independent random units as shown in Fig.~\ref{fig:reversmask}, and masking continuous regions. The detailed analysis refers to Sec.~\ref{sec:ablation}. 

\subsubsection{Setting the value of \textit{reg\_prob}.} In our implements, $reg\_prob=1$ is constant for fair comparison with part-based methods. Taking GaitGL as an example, it can be seen as mask-based with special dropping out certain regions of activations.

\subsubsection{Setting the value of $p$.} We investigate the area ratio for the masked region in the experiments. $p$ is applied to 0.5 as the optimal ratio, which means half of the feature map tends to be perturbed.

\subsubsection{Setting the value of $\alpha$ and $\beta$.} The ReverseMask is a flexible and general regularization method. By setting the different values of $\alpha$ and $\beta$, the networks regularized by ReverseMask could perform extremely differently in representation learning. We found that ReverseMask enables a network to capture fine-grained representations, like conventional part-based methods doing. The ReverseMask resembles DropBlock when $\alpha=1, \beta=0$. When $\alpha,\beta$ are sampled from the uniform distribution, information about the structure can still be sent to further extraction. In summary, there are two variants of ReverseMask according to the value of $\alpha,\beta$. We illustrate these two variants of ReverseMask in different colors as shown in figure \ref{fig:reversmask}.


\subsection{ReverseMask Feature Extraction}
\label{subsection:fe}

The conventional part-based methods utilize a shared convolutional layer to extract local representations. While GaitPart~\cite{gaitpart} utilized two-dimensional filters, and GaitGL~\cite{gaitgl} adopted three-dimensional convolutions which enhanced spatiotemporal feature learning. In other words, focal convolutions apply to each part with shared parameters. The receptive field of the focal convolution layer is restricted, which leads to the issue of \textit{boundary}. To alleviate such boundary weakness, our proposed ReverseMask randomly generates a pair of zero-masking features when $\alpha=1,\beta=0$. Therefore, the convolutions can still capture the fine-grained representations without boundary neglect. In our implements, we use identical 3D convolution to GaitGL\cite{gaitgl} for a fair comparison with part-based methods, which can design as
    \begin{equation}
     \mathit{F_{l}} = W(X^{m}_{l_{out}}) + W(X^{p}_{l_{out}})
    \end{equation}
    where $W(\cdot)$ means a 3D convolution operation, and $X^{m}_{l_{out}}$ and $X^{p}_{l_{out}}$ are the paired feature generated by Eqn.~\ref{eq:featmask} and Eqn.~\ref{eq:featpaired}. As we see, the ReverseMask Feature Extraction layer remains a full feature map for further layers, which brings about many advantages.

\subsection{Variants of Building Block}
\label{subsection:block}

\subsubsection{Plain ReverseMask Block} is a plain-like block similar to plain architecture as shown in Fig.~\ref{fig:reversmask}. The Plain ReverseMask Block is general. It can resemble DropBlock-like regularization when $\alpha=1, \beta=0$, which refers to dropping Plain-RMB for short. Also, it resembles scaling regularization when $\alpha,\beta$ sample from the uniform distribution,  which refers to scaling Plain-RMB for short. In our experiments, both scaling and dropping Plain-RMB are superior to DropBlock.   

\subsubsection{Inception-like ReverseMask Block} is designed for two reasons: (1) It is necessary to design such multi-branches architecture to establish a fair comparison with GaitGL. (2) The ReverseMask with different settings changes the distribution of the training set. Therefore such Inception-like ReverseMask Block can take advantage of multiple branches.  As shown in Fig.\ref{fig:ratio}(d), three representations learned from each branch are then aggregated by feature fusion module, which denotes as
    \begin{equation}
    \label{eq:fusion}
      \textbf{F} = {F_g + F_d + F_s} \qquad \qquad  \textbf{F} \in \mathbb{R}^{ b_l \times n_l \times c^{out}_{l} \times h_l \times w_l }
    \end{equation}
where $F_g, F_d,$ and $F_s$ represent features obtained from global, dropping and scaling branch, respectively. In addition, this strategy is used in every stage, excluding the last one. The final representations are concatenated in the final stage , which is commonly used in \cite{gaitgl,cstl,gln}. The fusion module is represented as:
    \begin{equation}
    \label{eq:fusion2}
      \textbf{F} = concat(F_g, F_d, F_s) \qquad \textbf{F} \in \mathbb{R}^{ b_l \times n_l \times c^{out}_{l} \times 3h_l \times w_l }
    \end{equation}

\section{Experiments}
The proposed method is evaluated on two popular public datasets: CASIA-B and OU-MVLP. CASIA-B is easy to evaluate the robustness to different variations, and OU-MVLP is the largest public gait dataset. Implementation details, results, comparisons, ablation study, and analysis are presented in the following part of this section.

\subsection{Settings}

\subsubsection{Datasets.}
\textit{CASIA-B}~\cite{casiab} consists of 124 subjects with ten sequences under 11 views, resulting in $124 \times 10 \times 11 = 13640$ gait sequences. Each subject walked ten times in three conditions, \ie six in normal (NM), two with a bag (BG),  two with a coat (CL). To compare fairly, we follow the experiment protocol in~\cite{aaai2019gaitset} which is widely employed by many other methods. Our experiments are conducted on three different configurations: Small-scale Training (ST), Medium-scale Training (MT), and Large-scale Training (LT). CASIA-B is split into a training set with 24 subjects and a test set with 100 subjects in ST configuration. MT configuration split CASIA-B to a training set with 62 subjects and a test set with 62 subjects. LT configuration has 74 subjects for training and 50 subjects for testing. Each subject's first four sequences (NM\#01-NM\#04) are put into the gallery set in the test phase. The remaining two sequences of NM, BG, and CL are in three different probe sets respectively to evaluate the robustness to different variations.

\textit{OU-MVLP}~\cite{oumvlp} was created by Osaka University and is the largest public gait database. It contains 10307 subjects, and each subject walks twice under 14 views. So, there are  $2 \times 14 = 28$ sequences for each subject. In our experiments, we follow the same protocol used in~\cite{aaai2019gaitset} also. That means 5153 subjects in the training set and the other 5154 subjects in the test set. In the test phase, sequence NM\#-01 is put into the gallery set, and sequence NM\#-00 is in the probe set.

\subsubsection{Implementation details:} \textbf{(1)} Human body silhouettes are aligned, cropped and resized to $64\times44$ by the preprocessing method~\cite{oumvlp}. The sequence length is 30 frames in the training phase, while the whole sequence is used in the test phase. \textbf{(2)} The separate Batch All \textit{(BA+)} triplet loss~\cite{triplet} is applied to train our model. The batch size \textit{(p,k)} is set up as (8, 16) for CASIA-B and (32, 8) for OU-MVLP, respectively. Margin $m$ for triplet loss is set to 0.2. \textbf{(3)} The pooling parameter $p$ is set to 6.5 for Generalized-Mean pooling~\cite{gem}.  \textbf{(4)} The iteration is set to 60K, 80K, and 80K for ST, MT, and LT, respectively, in CASIA-B training. In OU-MVLP training, the iteration is increased to 210K since there is more data in OU-MVLP. 
\textbf{(5)} Adam~\cite{adam} is employed as the optimizer, and the initial learning rate is $1e-4$ with weight decay $5e-4$. The learning rate reduces to $1e-5$ after 70K iterations for MT and LT, while the learning rate changes to $1e-5$ after 150K, then $5e-6$ after 200K iterations for OU-MLVP.
\textbf{(6)} For OU-MVLP, we doubled channel sizes of all convolution layers since OU-MVLP has more data than in CASIA-B.

\subsubsection{Evaluation metric} Rank-1 accuracy excluding identical-view sequences is taken as the same evaluation metric used in~\cite{aaai2019gaitset} and some other state-of-the-art methods. The metric has been widely employed to evaluate the performance of cross-view gait recognition.

\subsection{Comparisons with State-of-the-Art}

\begin{table*}[ht]
  \scriptsize
  \centering
   \caption{Rank-1 accuracy (\%) on CASIA-B under all view angles, different settings, and conditions, excluding identical-view cases.
}
  \vspace*{-1em}
    \label{tab:casiab}
  \resizebox{0.98\textwidth}{!}{
    \begin{tabular}{c|c|c|c|c|c|c|c|c|c|c|c|c|c|c}
    \toprule
    \multicolumn{3}{c|}{Gallery NM\#1-4}  &\multicolumn{12}{c}{$0^{\circ}$-$180^{\circ}$} \\
    \hline
    \multicolumn{3}{c|}{Probe}    & $0^{\circ}$     & $18^{\circ}$    & $36^{\circ}$    & $54^{\circ}$    & $72^{\circ}$    & $90^{\circ}$    & $108^{\circ}$   & $126^{\circ}$   & $144^{\circ}$   & $162^{\circ}$   & $180^{\circ}$  & Mean\\
    \midrule

    \multicolumn{1}{c|}{\multirow{9}[2]{*}{\textbf{ST (24)}}} & \multicolumn{1}{c|}{\multirow{3}[2]{*}{NM\#5-6}} & GaitSet~\cite{aaai2019gaitset} & 64.6  & 83.3  & 90.4  & 86.5  & 80.2  & 75.5  & 80.3  & 86.0  & 87.1  & 81.4  & 59.6  & 79.5   \\
\cline{3-15}          &       & \multicolumn{1}{c|}{GaitGL~\cite{gaitgl}} & 77.0  & 87.8  & 93.9  & \textbf{92.7} & 83.9  & 78.7  & 84.7  & 91.5  & 92.5  & 89.3  & \textbf{74.4} & 86.0   \\
\cline{3-15}          &       & Ours  & \textbf{78.1} & \textbf{89.2} & \textbf{95.3} & 92.6  & \textbf{83.9} & \textbf{79.7} & \textbf{85.2} & \textbf{91.8} & \textbf{93.2} & \textbf{89.6} & 73.9  & \textbf{86.6}  \\
\cline{2-15}          & \multicolumn{1}{c|}{\multirow{3}[2]{*}{BG\#1-2}} & GaitSet~\cite{aaai2019gaitset} & 55.8  & 70.5  & 76.9  & 75.5  & 69.7  & 63.4  & 68.0  & 75.8  & 76.2  & 70.7  & 52.5  & 68.6   \\
\cline{3-15}          &       & \multicolumn{1}{c|}{GaitGL~\cite{gaitgl}} & 68.1  & 81.2  & 87.7  & 84.9  & 76.3  & 70.5  & 76.1  & 84.5  & 87.0  & 83.6  & 65.0  & 78.6   \\
\cline{3-15}          &       & Ours  & \textbf{70.6} & \textbf{81.7} & \textbf{88.9} & \textbf{86.9} & \textbf{76.3} & \textbf{71.1} & \textbf{77.6} & \textbf{85.7} & \textbf{88.8} & \textbf{83.8} & \textbf{67.2} & \textbf{79.9}  \\
\cline{2-15}          & \multicolumn{1}{c|}{\multirow{3}[2]{*}{CL\#1-2}} & GaitSet~\cite{aaai2019gaitset} & 29.4  & 43.1  & 49.5  & 48.7  & 42.3  & 40.3  & 44.9  & 47.4  & 43.0  & 35.7  & 25.6  & 40.9   \\
\cline{3-15}          &       & \multicolumn{1}{c|}{GaitGL~\cite{gaitgl}} & 46.9  & 58.7  & 66.6  & 65.4  & 58.3  & 54.1  & 59.5  & 62.7  & 61.3  & 57.1  & 40.6  & 57.4   \\
\cline{3-15}          &       & Ours  & \textbf{50.2} & \textbf{65.4} & \textbf{70.8} & \textbf{69.0} & \textbf{63.0} & \textbf{58.0} & \textbf{63.3} & \textbf{67.6} & \textbf{66.2} & \textbf{61.6} & \textbf{43.2} & \textbf{61.7}  \\

    \hline
    \multicolumn{1}{c|}{\multirow{9}[2]{*}{\textbf{MT (62)}}} & \multicolumn{1}{c|}{\multirow{3}[2]{*}{NM\#5-6}} & GaitSet~\cite{aaai2019gaitset} & 86.8  & 95.2  & 98.0  & 94.5  & 91.5  & 89.1  & 91.1  & 95.0  & 97.4  & 93.7  & 80.2  & 92.0   \\
\cline{3-15}          &       & \multicolumn{1}{c|}{GaitGL~\cite{gaitgl}} & \textbf{93.9} & 97.6  & 98.8  & 97.3  & 95.2  & 92.7  & 95.6  & 98.1  & 98.5  & 96.5  & 91.2  & 95.9   \\
\cline{3-15}          &       & Ours  & 93.6  & \textbf{98.1} & \textbf{98.8} & \textbf{97.7} & \textbf{95.6} & \textbf{93.6} & \textbf{95.9} & \textbf{98.8} & \textbf{98.8} & \textbf{96.9} & \textbf{92.1} & \textbf{96.4}  \\
\cline{2-15}          & \multicolumn{1}{c|}{\multirow{3}[2]{*}{BG\#1-2}} & GaitSet~\cite{aaai2019gaitset} & 79.9  & 89.8  & 91.2  & 86.7  & 81.6  & 76.7  & 81.0  & 88.2  & 90.3  & 88.5  & 73.0  & 84.3   \\
\cline{3-15}          &       & \multicolumn{1}{c|}{GaitGL~\cite{gaitgl}} & 88.5  & 95.1  & 95.9  & 94.2  & 91.5  & 85.4  & 89.0  & 95.4  & 97.4  & 94.3  & 86.3  & 92.1   \\
\cline{3-15}          &       & Ours  & \textbf{89.6} & \textbf{95.5} & \textbf{96.8} & \textbf{95.5} & \textbf{92.2} & \textbf{87.0} & \textbf{90.9} & \textbf{95.5} & \textbf{98.2} & \textbf{94.6} & \textbf{87.1} & \textbf{93.0}  \\
\cline{2-15}          & \multicolumn{1}{c|}{\multirow{3}[2]{*}{CL\#1-2}} & GaitSet~\cite{aaai2019gaitset} & 52.0  & 66.0  & 72.8  & 69.3  & 63.1  & 61.2  & 63.5  & 66.5  & 67.5  & 60.0  & 45.9  & 62.5   \\
\cline{3-15}          &       & \multicolumn{1}{c|}{GaitGL~\cite{gaitgl}} & 70.7  & 83.2  & 87.1  & 84.7  & 78.2  & 71.3  & 78.0  & 83.7  & 83.6  & 77.1  & 63.1  & 78.3   \\
\cline{3-15}          &       & Ours  & \textbf{73.1} & \textbf{85.9} & \textbf{90.6} & \textbf{88.4} & \textbf{80.6} & \textbf{75.5} & \textbf{81.5} & \textbf{86.5} & \textbf{87.4} & \textbf{81.4} & \textbf{66.5} & \textbf{81.6}  \\
    \hline

    \multicolumn{1}{c|}{\multirow{15}[2]{*}{\textbf{LT (74)}}} & \multicolumn{1}{c|}{\multirow{5}[2]{*}{NM\#5-6}} & GaitSet~\cite{aaai2019gaitset} & 90.8  & 97.9  & 99.4  & 96.9  & 93.6  & 91.7  & 95.0  & 97.8  & 98.9  & 96.8  & 85.8  & 95.0   \\
\cline{3-15}          &       & GaitPart~\cite{gaitpart} & 94.1  & 98.6  & 99.3  & 98.5  & 94.0  & 92.3  & 95.9  & 98.4  & 99.2  & 97.8  & 90.4  & 96.2   \\
\cline{3-15}          &       & \multicolumn{1}{c|}{GaitGL~\cite{gaitgl}} & 96.0  & 98.3  & 99.0  & 97.9  & 96.9  & 95.4  & 97.0  & 98.9  & 99.3  & 98.8  & 94.0  & 97.4   \\
\cline{3-15}          &       & \multicolumn{1}{c|}{3DLocal~\cite{3dlocal}} & 96.0  & \textbf{99.0} & \textbf{99.5} & \textbf{98.9} & 97.1  & 94.2  & 96.3  & 99.0  & 98.8  & 98.5  & \textbf{95.2} & 97.5   \\
\cline{3-15}          &       & Ours  & \textbf{96.5} & 98.4  & 99.2  & 98.0  & \textbf{97.1} & \textbf{95.5} & \textbf{97.4} & \textbf{99.2} & \textbf{99.3} & \textbf{99.1} & 95.0  & \textbf{97.7}  \\
\cline{2-15}          & \multicolumn{1}{c|}{\multirow{5}[2]{*}{BG\#1-2}} & GaitSet~\cite{aaai2019gaitset} & 83.8  & 91.2  & 91.8  & 88.8  & 83.3  & 81.0  & 84.1  & 90.0  & 92.2  & 94.4  & 79.0  & 87.2   \\
\cline{3-15}          &       & GaitPart~\cite{gaitpart} & 89.1  & 94.8  & 96.7  & 95.1  & 88.3  & 84.9  & 89.0  & 93.5  & 96.1  & 93.8  & 85.8  & 91.5   \\
\cline{3-15}          &       & \multicolumn{1}{c|}{GaitGL~\cite{gaitgl}} & 92.6  & 96.6  & 96.8  & 95.5  & 93.5  & 89.3  & 92.2  & 96.5  & 98.2  & 96.9  & 91.5  & 94.5   \\
\cline{3-15}          &       & \multicolumn{1}{c|}{3DLocal~\cite{3dlocal}} & 92.9  & 95.9  & \textbf{97.8} & \textbf{96.2} & 93.0  & 87.8  & 92.7  & 96.3  & 97.9  & \textbf{98.0} & 88.5  & 94.3   \\
\cline{3-15}          &       & Ours  & \textbf{93.7} & \textbf{97.0} & 97.3  & 95.8  & \textbf{94.9} & \textbf{91.4} & \textbf{93.5} & \textbf{97.3} & \textbf{98.3} & 97.3  & \textbf{92.4} & \textbf{95.3}  \\
\cline{2-15}          & \multicolumn{1}{c|}{\multirow{5}[2]{*}{CL\#1-2}} & GaitSet~\cite{aaai2019gaitset} & 61.4  & 75.4  & 80.7  & 77.3  & 72.1  & 70.1  & 71.5  & 73.5  & 73.5  & 68.4  & 50.0  & 70.4   \\
\cline{3-15}          &       & GaitPart~\cite{gaitpart} & 70.7  & 85.5  & 86.9  & 83.3  & 77.1  & 72.5  & 76.9  & 82.2  & 83.8  & 80.2  & 66.5  & 78.7   \\
\cline{3-15}          &       & \multicolumn{1}{c|}{GaitGL~\cite{gaitgl}} & 76.6  & 90.0  & 90.3  & 87.1  & 84.5  & 79.0  & 84.1  & 87.0  & 87.3  & 84.4  & 69.5  & 83.6   \\
\cline{3-15}          &       & \multicolumn{1}{c|}{3DLocal~\cite{3dlocal}} & 78.2  & 90.2  & 92.0  & 87.1  & 83.0  & 76.8  & 83.1  & 86.6  & 86.8  & 84.1  & 70.9  & 83.7   \\
\cline{3-15}          &       & Ours  & \textbf{78.9} & \textbf{91.5} & \textbf{93.1} & \textbf{91.1} & \textbf{85.6} & \textbf{81.0} & \textbf{85.2} & \textbf{89.0} & \textbf{90.9} & \textbf{87.3} & \textbf{72.9} & \textbf{86.0}  \\

    \bottomrule
    \end{tabular}%
 }
\label{comparision_casia}
\vspace*{-1em}
\end{table*}%

\textbf{Results on CASIA-B.} Tab.~\ref{tab:casiab} shows our performance on CASIA-B, with three training configurations. With the large-scale configuration, we obtain 97.7\%, 95.3\%, and 86.0\% on rank-1 accuracy under NM, BG, and CL, respectively. The performance of our proposed method surpasses the classical method GaitSet~\cite{aaai2019gaitset} by a large margin. Compared with two typical part-based methods, GaitPart~\cite{gaitpart} and GaitGL~\cite{gaitgl}, our model outperforms GaitPart by 7.3\% under the most challenging condition of clothing (CL). At the same time, it outperforms the significant accuracy of GaitGL with 2.4\% under CL. The comparison with part-based methods replied on the coarse partition demonstrated Inception-like ReverseMask Block's effectiveness in enhancing the model by integrating structural and fine-grained representations. A recent part-based method, 3DLocal~\cite{3dlocal}, models the dynamic motion information on learned partition differently from previous pre-defined parts. However, the result illustrates that our model outperforms 3DLocal under all walking conditions significantly, which shows that random partition with simpleness can compete with adaptive region localization. We analyze the boosting performance coming from two aspects: (1) The proposed ReverseMask regularization injects noise into the feature map, which helps to prevent models from overfitting training data. Therefore, a better generalization performance is obtained in the test dataset. (2) random mask sampling considers full local representations from silhouettes, while the conventional part-based methods neglect characteristics around the gap between horizontal stripes. Therefore, the mask-based model should be superior to the previous part-based method depending on fixed horizontal stripes. 

\textbf{Results on OU-MVLP.} Tab.~\ref{tab:mvlp} lists the rank-1 accuracy of our model and other state-of-the-art methods on OU-MVLP. Our method achieves the best performance of 90.9\% on cross-view gait recognition. It noticed that 3DLocal obtains equal accuracy to our method when the invalid probe sequences are included, but it only demonstrates its performance of 96.5\% if the invalid probe sequences are excluded. Our method achieves 97.5\% rank-1 accuracy, a much better result.

\begin{table*}[htbp]
    \caption{Rank-1 accuracy (\%) on OUMVLP dataset under different view angles, excluding identical-view cases. The top eight rows and bottom six rows show the results with and without invalid probe sequences, respectively.}
  \centering
  \label{tab:mvlp}
  \vspace*{-1em}
  \resizebox{0.88\textwidth}{!}{
    \begin{tabular}{cc|c|c|c|c|c|c|c|c|c|c|c|c|c|c}
    \toprule
    \multirow{2}[2]{*}{\textbf{Method}} & \multicolumn{14}{|c|}{\textbf{Probe View}}                                                            & \multicolumn{1}{c}{\multirow{2}[2]{*}{\textbf{Mean}}}  \\
\cline{2-15}    \multicolumn{1}{c|}{} & $0^{\circ}$ & $15^{\circ}$ & $30^{\circ}$ & $45^{\circ}$ & $60^{\circ}$ & $75^{\circ}$ & $90^{\circ}$ & $180^{\circ}$ & $195^{\circ}$ & $210^{\circ}$ & $225^{\circ}$ & $240^{\circ}$ & $255^{\circ}$ & $270^{\circ}$ &   \\
    \midrule

    \multicolumn{1}{c|}{GaitSet~\cite{aaai2019gaitset}} & 79.3  & 87.9  & 90.0  & 90.1  & 88.0  & 88.7  & 87.7  & 81.8  & 86.5  & 89.0  & 89.2  & 87.2  & 87.6  & 86.2  & 87.1   \\
    \hline
    \multicolumn{1}{c|}{GaitPart~\cite{gaitpart}} & 82.6  & 88.9  & 90.8  & 91.0  & 89.7  & 89.9  & 89.5  & 85.2  & 88.1  & 90.0  & 90.1  & 89.0  & 89.1  & 88.2  & 88.7   \\
    \hline
    \multicolumn{1}{c|}{GLN~\cite{gln}}   & 83.8  & 90.0  & 91.0  & 91.2  & 90.3  & 90.0  & 89.4  & 85.3  & 89.1  & 90.5 & 90.6 & 89.6  & 89.3  & 88.5  & 89.2   \\
    \hline
    \multicolumn{1}{c|}{GaitKMM~\cite{Zhang_2021_CVPR}}   & 56.2  & 73.7  & 81.4  & 82.0  & 78.4  & 78.0  & 76.5  & 60.2  & 72.0  & 79.8  & 80.2  & 76.7  & 76.3  & 73.9  & 74.7  \\
    \hline
    \multicolumn{1}{c|}{GaitGL~\cite{gaitgl}} & 84.9  & 90.2  & 91.1  & 91.5  & 91.1  & 90.8  & 90.3  & 88.5  & 88.6  & 90.3  & 90.4  & 89.6  & 89.5  & 88.8  & 89.7  \\
    \hline
    \multicolumn{1}{c|}{CSTL~\cite{cstl}} & 87.1  & 91.0  & 91.5  & 91.8  & 90.6  & 90.8  & 90.6  & 89.4  & 90.2  & 90.5  & 90.7  & 89.8  & 90.0  & 89.4  & 90.2  \\    
    \hline
     \multicolumn{1}{c|}{3DLocal~\cite{3dlocal}}  & 86.1 & 91.2 & \textbf{92.6} & \textbf{92.9} & \textbf{92.2} & 91.3 & 91.1 & 86.9 & \textbf{90.8} & \textbf{92.2} & \textbf{92.3} & \textbf{91.3} & \textbf{91.1} & \textbf{90.2} & \textbf{90.9} \\
    \hline
     \multicolumn{1}{c|}{Ours}  & \textbf{87.9} & \textbf{91.5} & 91.7 & 92.0 & 92.0 & \textbf{91.6} & \textbf{91.3} & \textbf{90.7} & 90.3 & 90.9 & 91.1 & 90.8 & 90.5 & \textbf{90.2} & \textbf{90.9} \\
     
    \hline
    \\

    \hline
    \multicolumn{1}{c|}{GaitSet~\cite{aaai2019gaitset}} & 84.5  & 93.3  & 96.7  & 96.6  & 93.5  & 95.3  & 94.2  & 87.0  & 92.5  & 96.0  & 96.0  & 93.0  & 94.3  & 92.7  & 93.3   \\
    \hline
    \multicolumn{1}{c|}{GaitPart~\cite{gaitpart}} & 88.0  & 94.7  & 97.7  & 97.6  & 95.5  & 96.6  & 96.2  & 90.6  & 94.2  & 97.2  & 97.1  & 95.1  & 96.0  & 95.0  & 95.1  \\
    \hline
    \multicolumn{1}{c|}{GLN~\cite{gln}}   & 89.3  & 95.8  & 97.9  & 97.8  & 96.0  & 96.7  & 96.1  & 90.7  & 95.3  & 97.7 & 97.5 & 95.7  & 96.2  & 95.3  & 95.6   \\
    \hline
    \multicolumn{1}{c|}{GaitGL~\cite{gaitgl}} & 90.5  & 96.1  & 98.0  & 98.1  & 97.0  & 97.6  & 97.1  & 94.2  & 94.9  & 97.4  & 97.4  & 95.7  & 96.5  & 95.7  & 96.2  \\
    \hline
    \multicolumn{1}{c|}{3DLocal~\cite{3dlocal}} & - & - & - & - & - & - & - & - & - & - & - & - & - & - & 96.5 \\
    \hline
    \multicolumn{1}{c|}{Ours} & \textbf{93.7} & \textbf{97.5} & \textbf{98.6} & \textbf{98.8} & \textbf{98.0} & \textbf{98.5} & \textbf{98.2} & \textbf{96.5} & \textbf{96.7} & \textbf{98.2} & \textbf{98.1} & \textbf{97.1} & \textbf{97.6} & \textbf{97.2} & \textbf{97.5} \\

    \bottomrule
    \end{tabular}%
}

  \label{comparision_oumvlp}%
      \vspace*{-1em}
\end{table*}%

\subsection{Ablation study}
\label{sec:ablation}

\subsubsection{Masking strategies.} We study five variants of mask sampling strategies. The illustration and implements are detailed in supplementary materials. Those variants of sampling strategies achieve comparable performance, while different sampling strategies impact a lot on performance in other visual recognition tasks. We analyze that other visual tasks can take advantage of RGB modality, while silhouette-based gait recognition tends to capture static features and motion patterns from low-informative data.

\begin{figure}[t!]
\centering
\includegraphics[width=0.99\linewidth]{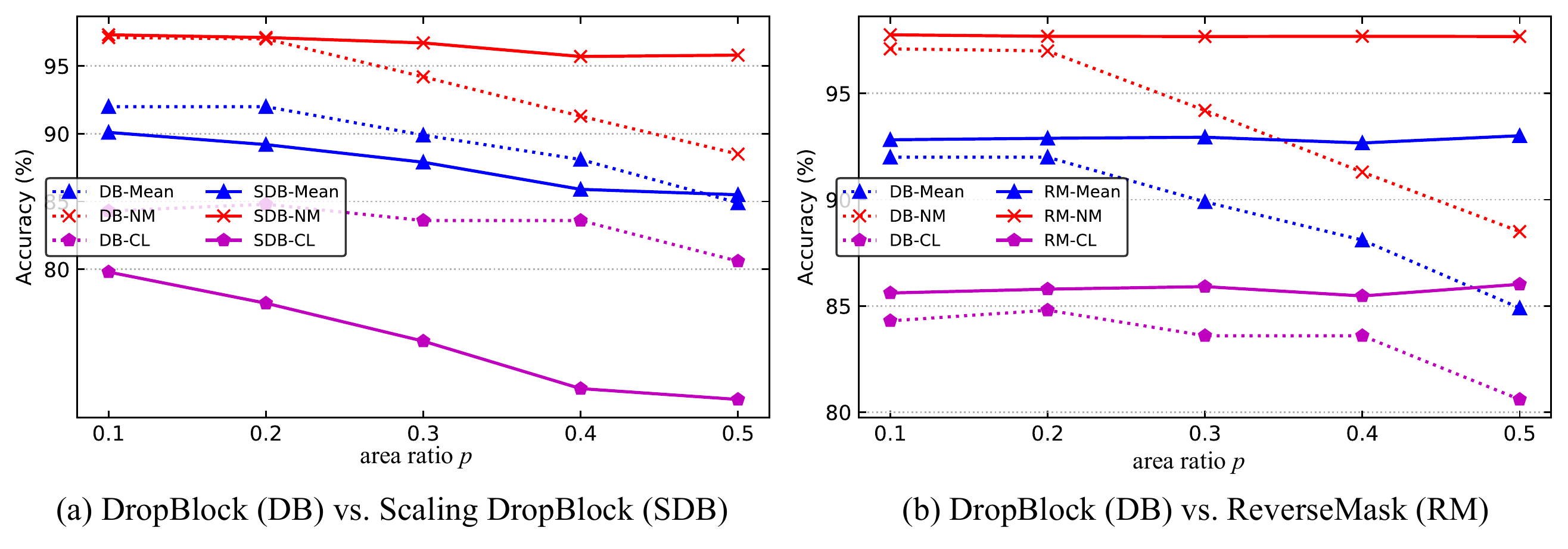}
\centering
\caption{
Performance comparison with different area ratio for sampling selected masking. 
}
\label{fig:comparisionblock}
\vspace{-5mm}
\end{figure}

\subsubsection{Impact of masked area ratio $p$.} As Fig.~\ref{fig:comparisionblock} shows, we found the stacked structure of DropBlock with the harsh setting of feature dropping can result in performance degradation. To our best knowledge, these results can be explained in two ways. (1) The harsh regularization changes data distribution dramatically; (2) The feature dropping in the shallow layer provides the incomplete feature map for the following layer, obstructing the following convolutions to learn representations. 

\subsubsection{Benefit of ReverseMask.} According to the analysis of performance degradation when harshly dropping feature map, we design ReverseMask Feature Extraction, which compensates the incomplete feature by its reversal masked feature. Albeit, the ReverseMask is simple, our experiments in Fig.~\ref{fig:comparisionblock}(b) show that it is effective to alleviate the issue of performance degradation led by conventional DropBlock.

\subsubsection{Mask-based vs. Part-based.} Our proposed mask-based regularizer can resemble part-based when the mask is constant instead of randomly generated. To illustrate the transition from mask-based model to conventional part-based model, we demonstrate such transition in the supplementary materials. The results in Tab.~\ref{tab:partbased} indicate that the mask-based model can outperform the part-based model by utilizing the information at the boundary, which is neglected by part-based methods.

\begin{table}[htbp]
\vspace*{-2em}
\caption{Gait recognition performance reported in \textit{rank-1} accuracy on CASIA-B.}
\vspace*{-1em}
\centering
\footnotesize
\resizebox{0.8\linewidth}{!}{
\begin{tabular}{l|l|lll|l}
 \hline
\textbf{Model}                                & Dim    & NM   & BG   & CL   & \textbf{Mean}  \\ \hline\hline
Baseline~\cite{gaitgl}                             & 4096   & 97.0 & 93.9 & 83.3 & 91.4 \\ \hline 
Baseline + part-based~\cite{gaitgl}                         & $4096\times2$ & 97.4 & 94.5 & 83.8 & 92.0 \\ \hline 
Baseline + Inception RMB(\textit{w/o} dropping branch) & $4096\times2$ & 97.6 & 94.8 & 85.1 & 92.5 \\ \hline
Baseline + Inception RMB(\textit{w/o} scaling branch)  & $4096\times2$ & 97.5 & 94.7 & 85.2 & 92.5 \\ \hline
Baseline + Inception RMB(\textit{w/o} global branch)   & $4096\times2$ & 97.6 & 95.0 & 85.1 & 92.6 \\ \hline
Baseline + Inception RMB                      & $4096\times3$ & \textbf{97.7} & \textbf{95.3} & \textbf{86.0} & \textbf{93.0} \\ \hline
\end{tabular}
}
\label{tab:partbased}
\vspace*{-2em}
\end{table}

\subsubsection{Feature scaling vs. Feature dropping.} As shown in Fig.\ref{fig:comparisionblock}(a), the curve reveals two significant phenomenon: (1) Although the trends of performance are the same for both scaling and dropping regularization, the model performs robust in the condition of clothing setting when dropping regularization is applied. (2) The model with scaling regularization obtains better accuracy on both normal and bag-carrying conditions. In the literature\cite{mesh,pami2020gaitnet}, gait structural representations contribute to distinguish subjects especially for normal and bag-carrying conditions, since the appearance information are relatively complete in these two conditions. It conducts that feature scaling regularizes model to learn structural representations which is robust on the conditions of normal wearing and bag-carrying, and feature dropping regularization enforces the networks to look at fine-grained features which is robust on the clothing condition.

\subsubsection{Variants of Building Block.} Two variants of blocks are built for evaluating proposed ReverseMask regularization method. As mentioned previously, Plain ReverseMask Block although has not improved performance to baseline model, but it helps model alleviate the drawback of DropBlock and prevents network from performance degradation. Besides, Scaling Plain ReverseMask Block achieves better performance than Dropping Plain ReverseMask Block. Notably, the most significant improvement is obtained when Inception-like ReverseMask Block is used to aggregate global, structural, and local representations. In addition, we think that the Plain ReverseMask Block is still with potential superiority to baseline model since the $reg\_prob$ set to one. In other words, it means this regularization configuration only supervises model to fix scaling data without any original silhouettes or features.

\subsubsection{Extend to other methods.} The proposed Inception-like ReverseMask Block is not only effective, but also plug-and-play. The ReverseMask regularization can generalize to the majority of gait methods, We study the extensive ReverseMask to three representative models. In Tab.\ref{fig:extend}, all models gain performance improvement after integrating ReverseMask regularizer to enhance discriminativeness of representations. We can observe that the model based on the Conv3D can gain relatively higher improvements. We analysis it is because Conv3D enhances models ability by superiority spatio-temporal representations learning.

\begin{table}[htbp]
\vspace*{-1.5em}
\caption{Effectiveness of ReverseMask regularization. To be noticed that the results of GaitSet are reproduced by~\cite{OpenGait}.}
\vspace*{-1em}
\centering
\footnotesize
\resizebox{0.80\linewidth}{!}{
\begin{tabular}{l|lll|l}
\hline
\textbf{Model}                & NM   & BG   & CL   & \textbf{Mean} \\ \hline
OpenGait                      & 96.3 & 92.2 & 77.6 & 88.7          \\ \hline
OpenGait + Inception-like ReverseMask Block & 97.0 (\textbf{0.7}) & 92.2 (0.0) & 79.4 (\textbf{1.8}) & 89.5 (\textbf{0.8})         \\ \hline
GaitSet                       & 95.9                                       & 91.3                                       & 74.8 &  87.3       \\ \hline
GaitSet + Inception-like ReverseMask Block   &  96.1 (\textbf{0.2}) & 91.3 (0.0)                        &  76.3 (\textbf{1.5}) & 87.9 (\textbf{0.6})         \\ \hline
Baseline                      & 97.0                                       & 93.9                                       & 83.3  & 91.4          \\ \hline
Baseline + Inception-like ReverseMask Block & 97.7 (\textbf{0.7}) &  95.3 (\textbf{1.4}) &  86.0 (\textbf{2.7}) & 93.0 (\textbf{1.6})        \\ \hline
\end{tabular}
}
\label{fig:extend}
\vspace*{-3em}
\end{table}

\subsubsection{Comparison to other regularization techniques.} We compare ReverseMask to random erasing and DropBlock, which are two commonly used regularization techniques. In Tab.~\ref{tab:augmentation}, Inception-like ReverseMask Block has better performance than not only feature-level but also input-level erasing regularization methods. Besides, we train baseline model with Inception-like ReverseMask Block and random erasing, and it achieves the best performance which shows our proposed feature-level regularization method can complementary to other regularization techniques.

\begin{table}[htbp]
\vspace*{-1.5em}
\caption{Performance of different regularization techniques. \textbf{RMB} donates ReverseMask Block for short. While Scaling DropBlock is a DropBlock-like regularization, Scaling DropBlock scales feature map rather than dropping. }
\vspace*{-1em}
\centering
\footnotesize
\resizebox{0.85\linewidth}{!}{
\begin{tabular}{l|lll|l}
\hline
\textbf{Model}                            & NM   & BG   & CL   & \textbf{Mean}  \\ \hline\hline
Baseline                         & 97.0 & 93.9 & 83.3 & 91.4 \\ \hline
Baseline + Random Erasing($reg\_prob$=0.5)       & 97.8 & 94.9 & 83.9 & 92.2 \\ \hline
Baseline + DropBlock($reg\_prob$=0.5,$p$=0.5)      & 96.2 & 93.3 & 84.5 & 91.3 \\ \hline
Baseline + DropBlock($reg\_prob$=1,$p$=0.5)              & 88.5 & 85.6 & 80.6 & 84.9 \\ \hline
Baseline + Scaling DropBlock($reg\_prob$=1,$p$=0.5)      & 95.8 & 90.3 & 70.4 & 85.5 \\ \hline 
Baseline + Dropping Plain RMB($reg\_prob$=1)             & 96.7 & 93.3 & 82.6 & 90.9 \\ \hline
Baseline + Scaling Plain RMB($reg\_prob$=1)              & 97.1 & 94.1 & 83.0 & 91.4 \\ \hline\hline
Baseline + Inception RMB($reg\_prob$=1)                  & 97.7 & 95.3 & 86.0 & 93.0 \\
Baseline + Inception RMB($reg\_prob$=1) + Random Erasing($reg\_prob$=0.5) & 98.1 & 96.0 & 86.9 & 93.7 \\ \hline
\end{tabular}
}
\label{tab:augmentation}
\vspace*{-3em}
\end{table}

\section{Conclusion}
    In this paper, we propose a novel mask-based regularization method called ReverseMask for obtaining better generalization performance, while it also vanishes the problems of overfitting and boundary existing in previous methods. To generalize the ReverseMask to the majority of works, we present the Inception-like ReverseMask Block which is able to capture more discriminative representations. In particular, feature scaling branch tends to extract structural information from silhouettes appearance, and feature dropping effectively utilizes local representations. Extensive experiments verify that the proposed regularization method achieves appealing performance on the CASIA-B and OU-MVLP. The ReverseMask feature is potentially extended to other tasks as an effective regularizer for better generalization of model.

\clearpage
%
%
\bibliographystyle{splncs04}
\bibliography{egbib}
\end{document}